\documentclass[letterpaper,10pt]{article} 
\usepackage{opticameet3} %% use version 3 for proper copyright statement

\newcommand\authormark[1]{\textsuperscript{#1}}

\usepackage{amsmath,amssymb}
\usepackage[colorlinks=true,bookmarks=false,citecolor=blue,urlcolor=blue]{hyperref} %pdflatex
\usepackage[caption=false,font=footnotesize]{subfig}
\usepackage[percent]{overpic}

\usepackage{url}

\def\j{\mathrm{j}}

\begin{document}

% \title{Neural Probabilistic Amplitude Shaping for Nonlinearity-Tolerant Optical Transmission}
\title{Neural Probabilistic Amplitude Shaping for Nonlinear Fiber Channels
}

% \author{Author name(s)}
% \address{Author affiliation and full address}
% \email{e-mail address}

\author{Mohammad Taha Askari,\authormark{1,2,*} Lutz Lampe,\authormark{1} and Amirhossein Ghazisaeidi\authormark{2}}

\address{\authormark{1}Department of Electrical and Computer Eng., University of British Columbia, Vancouver, BC V6T 1Z4, Canada\\
\authormark{2}Nokia Bell Labs, 12 rue Jean Bart, 91300 Massy, France}

\email{\authormark{*}mohammadtaha@ece.ubc.ca} %% email address is required

%% Do not add a copyright statement. Optica will add it.

%\begin{abstract}
%We introduce neural probabilistic amplitude shaping (NPAS), a joint-distribution learning framework for probabilistic amplitude shaping. The proposed scheme provides a 0.1~bits/2D achievable information rate gain for dual-polarized 64-QAM transmission over a single-span 205~km link.
%\end{abstract}

% % mentions SNR gain - compares with sequence selection - no repetetion of "probabilsitic amplitude shaping"
 \begin{abstract}
We introduce neural probabilistic amplitude shaping, a joint-distribution learning framework for coherent fiber systems. The proposed scheme provides a 0.5~dB signal-to-noise ratio gain over sequence selection for dual-polarized 64-QAM transmission across a single-span 205~km link.
 \end{abstract}

\section{Introduction}
Probabilistic amplitude shaping (PAS) is widely used in coherent optical systems because it enables probabilistic shaping (PS) and fine rate adaptation while remaining fully compatible with systematic forward error correction (FEC) \cite{bocherer2015bandwidth}. However, traditional PAS shapes only marginal symbol probabilities, even though nonlinear fiber channels are sensitive to both individual symbol distributions and their temporal dependencies. This gap motivates shaping techniques that optimize the joint distribution of symbol sequences rather than treating symbols independently.

Finite-blocklength PAS can inherently introduce beneficial  symbol correlations that reduce nonlinear interference noise (NLIN) and increase the achievable information rate (AIR) \cite{fehenberger2020analysis_short}. Sequence selection methods  exploit this by generating  multiple candidate sequences with a fixed marginal distribution and selecting one via a nonlinear metric \cite{civelli2024sequence, askari2024probabilistic}. This approach indirectly optimizes the joint distribution of transmitted symbols, but its practicality is limited: the method requires a large candidate pool to achieve notable gain, incurs high computational complexity and rate loss, and offers no control over candidate quality \cite{civelli2024cost}. These drawbacks have motivated data-driven alternatives that directly learn useful temporal dependencies.

Neural networks have been applied to optimize PS in an end-to-end learning framework, where a neural model parameterizes the symbol distribution and is trained over a differentiable channel using information-theoretic objectives such as bit-metric information or its proxy, the binary cross-entropy (BCE) loss \cite{ ait2020joint, chimmalgi2025end}. Early neural shaping methods, however, still focus solely on marginal symbol probabilities.

More recently, neural PS (NPS) has been proposed as a general framework for learning the joint distribution of transmitted symbols using autoregressive recurrent neural networks \cite{askari2025neuralECOC}. By capturing temporal dependencies, NPS outperforms sequence selection methods and effectively mitigates NLIN. However, because it operates on signed symbols, NPS cannot  integrate with PAS or systematic FEC and thus remains primarily a performance benchmark rather than a deployable shaping scheme.

In this work, we introduce neural PAS (NPAS), a new PAS-compatible joint-distribution learning method for unsigned symbols. NPAS employs an autoregressive recurrent neural network trained end-to-end through a differentiable optical-fiber channel to model the joint amplitude distribution. By constraining learning to the PAS-relevant portion of the constellation, NPAS retains the benefits of joint-distribution shaping while ensuring interoperability with systematic FEC and standard PAS architectures. Furthermore, the reduced optimization space markedly improves training stability at high-dimensional settings. We show that NPAS achieves shaping and nonlinear-mitigation gains on par with NPS, while being directly deployable in practical PAS transceivers.

\section{NPAS Architecture and Training}
\label{sec:architecture}
NPAS consists of an autoregressive neural model that generates a sequence of conditional distributions over unsigned complex symbols. During operation, an arithmetic distribution matcher (ADM) \cite{baur2015arithmetic} uses these  distributions to deterministically map information bits to unsigned symbols. After each symbol is generated, the neural model updates its internal state, producing the next conditional distribution in the sequence. The corresponding sign bits are supplied by the FEC parity stream, ensuring full compatibility with the PAS architecture.

\begin{figure}[t]
  \centering
  \includegraphics[width=\textwidth]{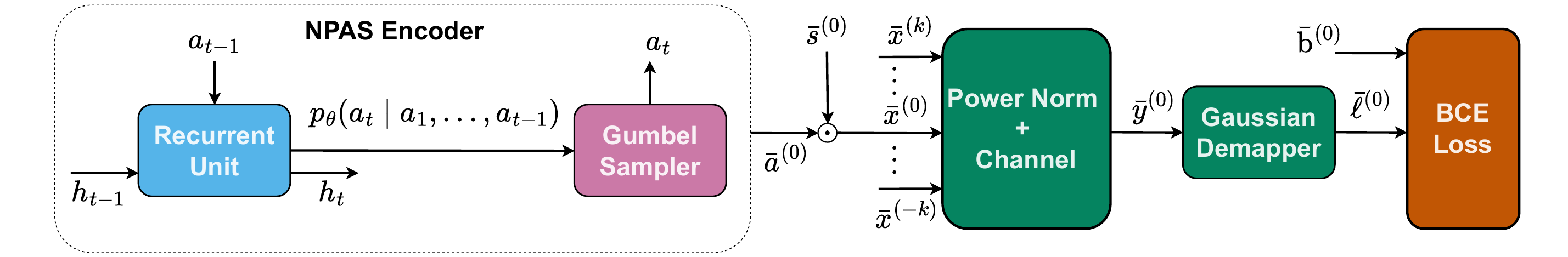}
  \vspace{-5mm}
  \caption{NPAS end-to-end training pipeline.}
  \label{fig:npas-blk}
  \vspace{-5mm}
\end{figure}

During training, the autoregressive model is learned end-to-end so that its conditional distributions become optimal for nonlinear fiber propagation. The training pipeline is shown in Fig.~\ref{fig:npas-blk}. The NPAS encoder models the $L$-dimensional joint distribution over unsigned complex symbols $\bar{a}=(a_1,\dots,a_L)$ as $p_\theta(\bar{a}) = \prod_{t=1}^{L} p_\theta(a_t \mid a_1,\dots,a_{t-1})$,
where each conditional term is parameterized by an autoregressive neural model $\theta$. At each time step, a recurrent unit generates logits representing the conditional probability of the next unsigned symbol given the previous unsigned symbol $a_{t-1}$ and the recurrent context vector $h_{t-1}$ capturing the temporal dependencies among past amplitudes. During training, discrete samples are drawn using the Gumbel-Softmax trick \cite{jang2016categorical} with a straight-through estimator \cite{bengio2013estimating}.
This procedure is repeated $L$ times to sample $\bar{a}$ from the learned joint distribution, while the symbol signs $\bar{s}=(s_1,\dots,s_L)$, where each $s_i$ is an in-phase and quadrature sign pair drawn independently from a uniform Bernoulli distribution, are applied to form the sequence of signed constellation points $\bar{x}=(x_1,\dots,x_L)$. To incorporate channel memory effects, the main sequence $\bar{x}^{(0)}$ is concatenated with its neighboring blocks $\bar{x}^{(-k)},\dots,\bar{x}^{(k)}$. The extended sequence is normalized to enforce a transmit power constraint and then passed through a differentiable single-channel single-polarization fiber model.

To account for nonlinear fiber effects during training, we adopt the additive-multiplicative (AM) perturbative model \cite{askari2024probabilistic}, which represents the received symbols as
\vspace{-1mm}
\begin{equation}
\label{eq:channel}
y_t = x_t \exp\!\left(\j \gamma \sum_{n} (|x_{t-n}|^2 - 1)c_n \right) + \Delta x_t + n_t ,
\end{equation}
where $\gamma$ is the nonlinear coefficient, $c_n$ are multiplicative perturbation coefficients, $\Delta x_t$ represents additive nonlinear interference, and $n_t$ denotes amplified spontaneous emission (ASE) noise. This model captures first-order nonlinear phase rotation and additive Kerr-induced distortions in the presence of a constant carrier-phase recovery (CPR) while remaining differentiable, making it well suited for gradient-based end-to-end training.
After propagation, the central block $\bar{y}^{(0)}$ is extracted and processed by a mismatched Gaussian demapper to compute log-likelihood ratios $\bar{\ell}^{(0)}$, which are compared with the transmitted bit labels $\bar{b}^{(0)}$ using the adjusted BCE loss from \cite{askari2025neuralECOC}. This loss serves as a differentiable surrogate for maximizing AIR, enabling the NPAS encoder to learn temporal amplitude dependencies that are jointly optimal for nonlinear fiber propagation.

%------------------------------------------ Results --------------------------------------------%
\section{Numerical Results}
We adopt the simulation setup from \cite{askari2025neuralECOC, gultekin2022kurtosis}: a 205~km single-span single-mode fiber link carrying 5 wavelength-division multiplexing (WDM) dual-polarization 64-QAM  channels at 50~GBd (55~GHz spacing),  with 17~ps/nm/km chromatic dispersion (CD), nonlinear coefficient $\gamma = 1.3~\mathrm{W^{-1}km^{-1}}$, 0.2~dB/km attenuation, and post-span  erbium-doped fiber amplifier (EDFA) with 5~dB noise figure. Root-raised-cosine (roll-off 0.1) pulse shaping and  electronic CD compensation and pilot-aided linear CPR with $2.5\%$ pilot rate \cite{neshaastegaran2019log} are applied.

The NPAS encoder employs a single-layer long short-term memory (LSTM) \cite{hochreiter1997long} with hidden size 256, followed by a projection layer mapping the hidden state to  unsigned constellation dimension. For a fair comparison, the same architecture is used for NPS, with a mapping to signed symbols. For each blocklength $L$, we choose the number of neighboring blocks $k$ such that the concatenated side blocks cover the effective one-sided channel memory introduced by nonlinear propagation. The model is then trained at its optimal launch power (see Section~\ref{sec:architecture}) and evaluate the generated symbols in the WDM setup over varied launch powers using split-step Fourier method (SSFM)  to faithfully emulate physical fiber propagation.

\begin{figure*}[t!]
\centering
\includegraphics[width=\textwidth]{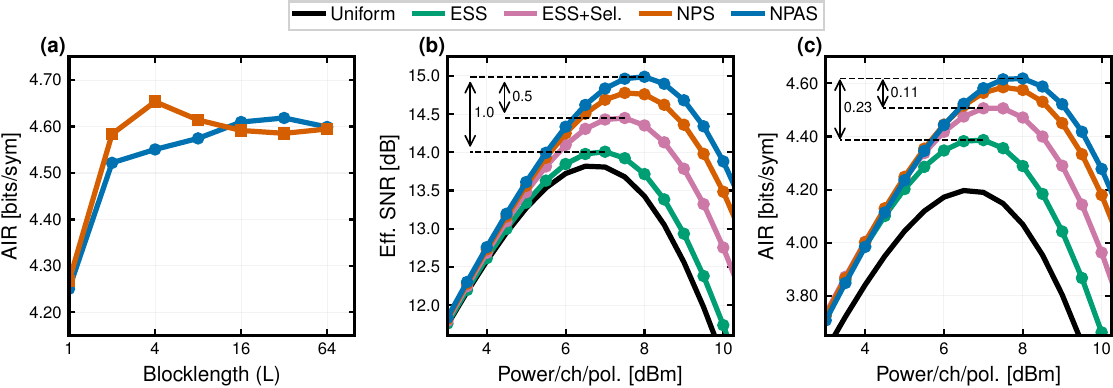}
\caption{Performance comparison of NPAS with conventional PAS using ESS and with NPS. (a) AIR vs.\ blocklength $L$; (b) effective SNR vs.\ launch power; and (c) AIR vs.\ launch power.}
\label{fig:air_snr_vs_power}
\vspace*{-5mm}
\end{figure*}

Fig.~\ref{fig:air_snr_vs_power}(a) shows AIR versus blocklength $L$ for NPAS, with NPS included for comparison. As $L$ grows, learning shifts from marginal to joint distributions, letting both methods exploit temporal dependencies. For $L\le 8$, NPS edges out NPAS by directly optimizing signed symbols. Beyond that, NPS plateaus and then degrades, while the proposed NPAS remains stable and continues to improve. This effect is attributed to the $4^L$ times smaller search space of NPAS. Thus, NPAS incurs a small penalty at low dimensions but maintains stability at higher dimension, achieving NPS-level performance while retaining deployability.

Figs.~\ref{fig:air_snr_vs_power}(b) and (c) compare NPAS against uniform signaling and conventional PAS with enumerative sphere shaping (ESS) \cite{amari2019introducing-short}, both with and without sequence selection. We use blocklength $L=32$ as a practical operating point that balances low rate loss with manageable computational complexity. We set the ESS shaping rate to $1.93$~bits/1D to match the entropy of the NPS marginal ($L=1$) case, and map amplitudes to the in-phase and quadrature components of the QAM symbols as described in \cite{askari2023probabilistic}. Sequence selection draws from $64$ candidates generated through interleaving and scored by the AM-based metric from \cite{askari2023probabilistic}.  NPS is included for reference. Because NPAS and NPS are trained via Gumbel-Softmax sampling, the ADM rate loss is not yet quantified; therefore, we leave ESS's rate losses, including those from sequence selection, undeducted to provide an optimistic AIR bound for ESS as our benchmark. At optimal launch power, NPAS and NPS outperform ESS and ESS+selection by more than $0.5$~dB in effective SNR and $0.1$~bits/2D in AIR, and their  optimal launch power is $\approx0.5$~dB higher. This shift confirms superior nonlinear-interference mitigation via joint-distribution learning. These results show that direct joint-distribution learning beats candidate-based sequence selection, while NPAS alone preserves full PAS compatibility without incurring selection-related  complexity and rate penalties.

%------------------------------------------ Conclusion --------------------------------------------%
\section{Conclusion}
We introduced neural probabilistic amplitude shaping (NPAS), a PAS-compatible neural joint-distribution learning method for optical fiber systems. NPAS learns the joint distribution over unsigned symbols and therefore integrates seamlessly with PAS while retaining its shaping benefits. The results show that NPAS achieves comparable gain to NPS, while offering markedly superior stability for larger blocklengths due to its reduced optimization space. These findings confirm that joint-distribution learning can be realized within a PAS architecture without sacrificing practicality, making NPAS a promising candidate for next-generation coherent transceivers. The results presented here are limited to a single-span link, and extending the comparison to multi-span transmission is subject of future 
investigation.

\section{Acknowledgements}
This work was supported by Nokia Bell Labs (France), Mitacs Accelerate International, UBC's Institute for Computing, Information and Cognitive Systems, and the Digital Research Alliance of Canada (www.alliancecan.ca).

\bibliographystyle{opticajnl}
\bibliography{sample}

\end{document}